\documentclass[lettersize,journal]{IEEEtran}
\usepackage{amsmath,amsfonts}
\usepackage{algorithmic}
\usepackage{algorithm}
\usepackage{array}
\usepackage[caption=false,font=normalsize,labelfont=sf,textfont=sf]{subfig}
\usepackage{textcomp}
\usepackage{stfloats}
\usepackage{multirow}
\usepackage{url}
\usepackage{verbatim}
\usepackage{graphicx}
\usepackage{float}
\usepackage{caption}
\usepackage{tabularx}
\usepackage{array}
\usepackage{booktabs}

\newcolumntype{C}{>{\centering\arraybackslash}p{3cm}}

\begin{document}

\title{Exploring Accurate 3D Phenotyping in Greenhouse through Neural Radiance Fields}

\author{Junhong Zhao$^{1,*}$, Wei Ying$^{2,*}$, Yaoqiang Pan$^{2}$, Zhenfeng Yi$^{1}$, Chao Chen$^{3}$, Kewei Hu$^{2,\#}$, Hanwen Kang$^{2,\#}$ \\[2pt]
${*}$ \small{Equal contribution} \\
${^1}$ \small{Guangdong Academic of Agriculture Science} \\
${^2}$ \small{College of Engineering, South China Agriculture University} \\
${^3}$ \small{Department of Mechanical and Aerospace Engineering, Monash University}
}

\twocolumn[{
\renewcommand\twocolumn[1][]{#1}
\begin{center}
    \maketitle
    \captionsetup{type=figure}
    \includegraphics[width=1\textwidth]{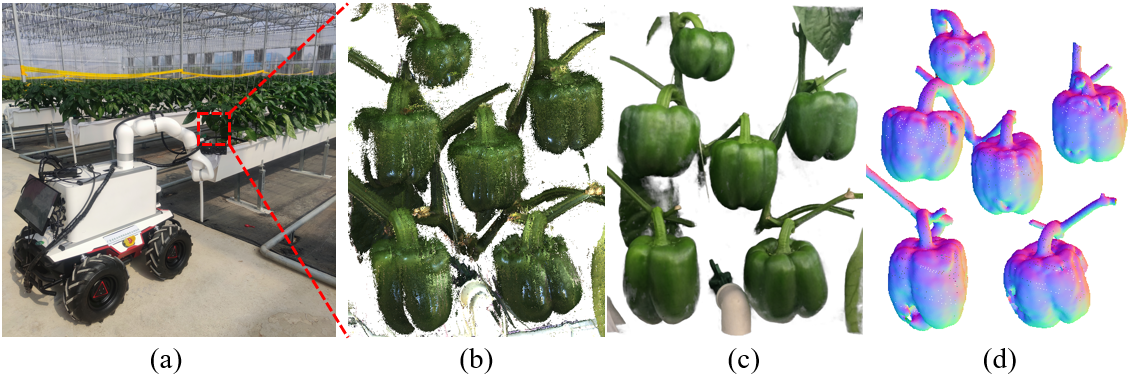}
    \captionof{figure}{Phenotyping results by respectively using the 3D scanner and NeRF reconstruction in the greenhouse by robots.}
    \label{fig:1}
\end{center} }]

\begin{abstract}
Accurate collection of plant phenotyping is critical to optimising sustainable farming practices in precision agriculture. 
Traditional phenotyping in controlled laboratory environments, while valuable, falls short in understanding plant growth under real-world conditions. 
Emerging sensor and digital technologies offer a promising approach for direct phenotyping of plants in farm environments. 
This study investigates a learning-based phenotyping method using the Neural Radiance Field to achieve accurate in-situ phenotyping of pepper plants in greenhouse environments. 
To quantitatively evaluate the performance of this method, traditional point cloud registration on 3D scanning data is implemented for comparison. 
Experimental result shows that NeRF(Neural Radiance Fields) achieves competitive accuracy compared to the 3D scanning methods.
The mean distance error between the scanner-based method and the NeRF-based method is 0.865mm.
This study shows that the learning-based NeRF method achieves similar accuracy to 3D scanning-based methods but with improved scalability and robustness.
\end{abstract}

\section{Introduction} \label{section: Introduction}
The development of artificial intelligence and high-precision sensors has changed the paradigm of agricultural production. 
Precision agriculture is rapidly evolving, significantly improving efficiency and productivity compared to traditional production methods \cite{sishodia2020applications}. 
In precision agriculture, the measurement of plant phenotypes is of critical importance. 
Plant phenotyping is an evolving science that follows plant genomics and ecophysiology \cite{feng2021comprehensive, fu2020application}. 
Phenotyping provides a rapid understanding of the traits expressed by genes with complex structures and helps to understand genetic characteristics for different plant functions. 
In addition, plant breeders require continuous growth experiments to select the optimal environment for plant growth. 
Phenotyping is a complex and challenging task due to the numerous characteristics of genotypes \cite{schauer2006plant}\cite{schulze2010quantitation}. 
Historically, plant breeders have performed manual phenotyping, which is an expensive, labour-intensive and time-consuming process. 
Therefore, there is a current need for easier and more accurate phenotyping measurements \cite{meraj2024computer}.

2D imaging can be achieved with an RGB camera to measure basic morphological features of the plant such as colour, shape and texture.
However, the geometric appearance is missing because the data is limited to two dimensions \cite{zhang2018imaging}. 
3D imaging systems can capture basic geometric features of plants, such as volume, stem angle and projected canopy area, with high quality. 
In addition, 3D methods can track plant growth and yield over time, helping researchers to make phenotyping judgments \cite{lou2015estimation}\cite{kolhar2023plant}. 
High-quality 3D reconstruction models can be used to characterise leaves, estimate crop yields and classify fruits \cite{paturkar20193d}\cite{harandi2023make}.
In recent years, several technological approaches have been developed to quickly and accurately acquire three-dimensional data on plant morphology and structure. 
These approaches include the use of instruments such as depth cameras \cite{mccormick20163d} and 3D scanners \cite{wu2019accurate} to obtain high-quality 3D point cloud data of fruit \cite{wen2024accurate}. 
Multi-View Stereo (MVS) is a more efficient method than 3D scanning for phenotype acquisition using multiple cameras \cite{wu2020mvs}. NeRF, a neural network-based approach for Novel View Synthesis, enables rapid reconstruction of 3D models by learning information from 2D images \cite{mildenhall2021nerf}. In various scenarios, NeRF can offer valuable insights that are hard to obtain using 2D data alone. This is achieved by incorporating multi-view data, which helps to overcome the limitations caused by occlusions and crossings in plant structures. NeRF can reconstruct the distance, orientation, and light of the plant, providing new perspectives. Despite being an implicit field representation, NeRF stores density information in the neural network, which serves as a crucial database for later geometric extraction.

Although there have been advancements, challenges still exist with the current method of acquiring phenotyping data. 
Firstly, using a high-precision 3D scanner is expensive and requires specific acquisition environments. 
Secondly, processing the acquired point cloud data is laborious, involving multiple steps such as filtering and alignment due to the dense and continuous nature of the point cloud.  
As a result, this leads to significant modelling time.
In addition, 3D reconstruction methods based on MVS require high-resolution and precise measurements, as well as continuous acquisition of multi-view images. 
The strict equipment requirements lead to longer processes and reduced overall robustness.
Although current 3D phenotype reconstruction methods that use high-fidelity neural radiation fields can produce high-quality geometries, there is still a gap in recovering the actual dimensions of the scene for real-data measurements of the model.

Therefore, this study investigates both traditional 3D scanner measurements and neural network-based NeRF 3D reconstruction methods. 
The high-precision 3D scanners used in this study provided the data necessary for a thorough evaluation of reconstruction quality. 
A comparative analysis was conducted on the quality of geometric models generated by various reconstruction methods, including scale restoration.
Specifically, the contributions of this paper are as follows:
\begin{itemize}
    \item A method for recovering the true scale of NeRF is introduced, which is compared with the high-precision fusion results of the point cloud to compute the true size.    
    \item Improved generalization ability and robustness of NeRF models in plant scenarios.
    \item A 3D semantic segmentation network was introduced to improve the accuracy of phenotype detection.
\end{itemize}

The rest of this paper is organised as follows. Section \ref{section: review} surveys related work. 
Section \ref{section:method} details the methodology for comparing point cloud measurements and NeRF reconstruction and the scale restoration.  
\ref{section:experiment} is a discussion of the results and details of the experiments. The conclusion in Section \ref{section:conclusion}.

\section{Related Works} \label{section: review}
\subsection{Plant Phenotyping}
Phenomics is an emerging field of research that quantifies animal and plant traits in multiple dimensions. It provides comprehensive scientific knowledge that is no longer limited to the study of a single trait \cite{zhang2023high}.
Traditional methods rely on imaging and 3D distance sensors to measure various plant traits such as colour, shape, volume and spatial structure \cite{li2020review, feng2021comprehensive}.
Han et al. used RGB cameras for fast and inexpensive data acquisition. However, the dimensionality of the measurements was limited to two dimensions, and high-dimensional data were lacking \cite{han2021rgb}.
In contrast, Zhu et al. used a depth camera (RGB-D) to obtain full-width image information of tomato canopies \cite{zhu2023method}.
Forero et al. scanned plant seedlings using 3D LiDAR to obtain point clouds of the growth process \cite{forero2022lidar}.
However, while conventional sensor-based phenotyping devices are capable of collecting various types of phenotypic data, they are limited by their susceptibility to environmental factors and their demanding nature. 

\subsection{Explicit Modelling Approaches}
Traditional phenotypic model reconstruction methods typically involve explicit model reconstruction, with data storage methods including point clouds, meshes and voxels. 
Jay et al. used a digital camera to capture continuous images and incorporated a structure-from-motion (SFM) approach to reconstruct field crop models \cite{jay2015field}.
Kang et al. proposed a visual sensing and perception strategy based on LiDAR colour fusion for accurate scene understanding and fruit localisation in orchards \cite{kang2022accurate, kang2023semantic}.
Guo et al. used the Realsense depth camera to acquire multi-view images with depth information \cite{guo2023improved}.
Yang et al. used a UAV to acquire images with high-precision location information \cite{yang2023new}.
However, all of them require the integration of high-precision position information with image data to ensure model stability. 
In addition, a large number of images must be acquired continuously, and the reconstruction process is time-consuming, taking at least 25 minutes to achieve satisfactory results.

\subsection{Implicit Modelling Approaches}
The technique of encoding spatial scenes in a neural network by learning continuous mathematical functions is called implicit reconstruction. 
Implicit methods are particularly adept at describing complex topologies and continuous surfaces. 
NeRF-based approaches, such as those of Mildenhall et al, achieve accurate scene density values through basic ray tracing and volume rendering \cite{mildenhall2021nerf}.
While the density field reconstruction method can achieve high rendering quality, it may lack high-precision surface information. 
To address this limitation, Wang et al. proposed the Signed Distance Function (SDF) to improve the quality of surface reconstruction by imposing model constraints \cite{wang2021neus}.
In addition, Muller et al. introduced a multi-resolution hash table and voxel representation to significantly improve reconstruction speed \cite{muller2022instant}. 
However, the large weights used for model training often require image compression during the training phase. 
Consequently, this compression can lead to the loss of the true size of the model after reconstruction, making phenotypic measurements impossible.

Our research addresses this challenge by investigating NeRF for both appearance rendering and 3D modelling, effectively mitigating the scale discrepancy present in NeRF-based phenotype measurement methods. 
In addition, we quantitatively assess the accuracy of NeRF modelling in phenotyping by comparing it to high-precision point cloud registration results, an endeavour that, to the best of our knowledge, has not been done by previous work.

\newpage

\section{Methodologies} \label{section:method}
\begin{figure*}[ht]
    \centering
    \includegraphics[width=1\linewidth]{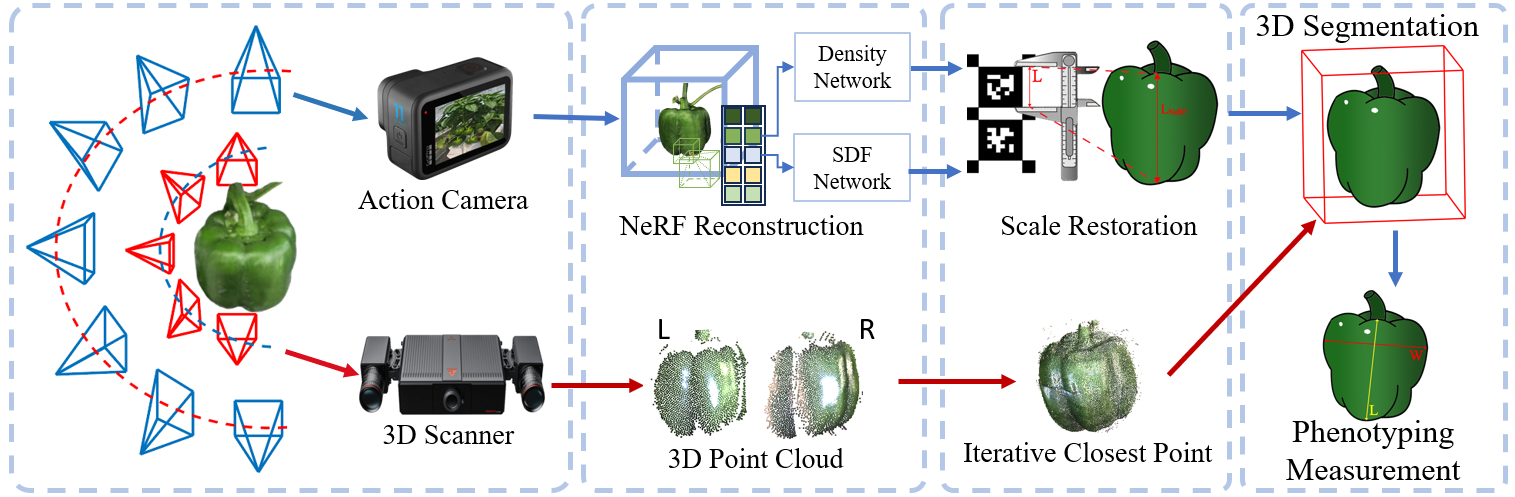}
    \caption{NeRF Scale Restoration Process Pipeline.}
    \label{fig:pipeline}
\end{figure*}
\subsection{Data Acquisition System}
This study uses two approaches to data collection. 
The first method utilises traditional image capture techniques using a high frame rate and high-resolution action camera, specifically the GoPro Hero 11. 
This camera is capable of capturing 4K resolution images at a frame rate of 120Hz, ensuring both quality and continuity in the data collection process. 
However, due to the limited view of plant occlusion within the scene, only one side of the plant can be captured. 
The chosen perspective is a frontal view of the area of interest, which requires precise sampling angles and equipment stability. 

\begin{table}[h]
    \center
    \caption{Technical Parameters of the Devices}
    \renewcommand{\arraystretch}{1.3}
    \footnotesize
    \label{table:1}
    \begin{tabular}{l|l|c}
        \toprule[2pt]
        \textbf{Devices} & \textbf{Parameter} & \textbf{Value} \\ \hline
        \multirow{4}{4em}{\textbf{GoPro HERO11}} & Weight & 149.00g \\
        ~ & Resolution & 4K+ \\
        ~ & Lens Stabilization & Electronic \\
        ~ & Battery Life & 90 minutes \\ \hline
        \multirow{10}{4em}{\textbf{RVC-X mini}} & 3D point cloud acquisition time & 1.5 \\
        ~ & Resolution & 1.6MP \\
        ~ & Recommended working distance(mm) & 250-1500 \\
        ~ & Camera Weight(kg) & 1.7 \\
        ~ & Camera Size(mm) & 286*110*50 \\
        ~ & Depth of field(mm) & 100 \\
        ~ & FOV(mm) & 147*116 \\
        ~ & XY-axis Resolution(mm) & 0.12 \\
        ~ & Z-axis repeatability(mm) & 0.03 \\
        ~ & Z-axis measurement accuracy(mm) & 0.003 \\
        \bottomrule[2pt]
    \end{tabular}
\end{table}

Secondly, high-precision structured light was used to acquire point cloud data from the image. 
The robotic arm used for this is the xArm6, while the structured light 3D point cloud scanner is the RVC-X mini. 
The robotic arm equipped with the 3D scanner is programmed to follow a path around the target to be measured. 
At nine predetermined positions along this path, the arm stops, allowing the 3D scanner to perform a point cloud scan and document the current position of the arm, including its coordinates in $x, y, z$ space and its rotation angles $Rx, Ry, Rz$.

\subsection{3D modelling from NeRF}
\subsubsection{Reconstruction from Neural Radiance Fields}
The traditional NeRF is a 3D reconstruction method that is based on ray tracing and uses the MLP as an implicit representation. Input the point position coordinates ($x, y, z$) and the viewing direction ($\theta,\phi$) into the ray equation $r(t)=o+t \mathbf{d}$, where $o$ is the position coordinate, $\mathbf{d}$ is the viewing direction, $t$ is the distance from the sampling point to the origin. In particular, the position coordinates and viewing direction need to be added to the encoding, and the encoding function is:
\begin{equation}
    \gamma(p)=\prod_{L=0}^{10}\left(\sin2^{L}\pi p\cdot\cos2^{L}\pi p\right).
\end{equation}

\begin{figure}[ht]
    \centering
    \includegraphics[width=1\linewidth]{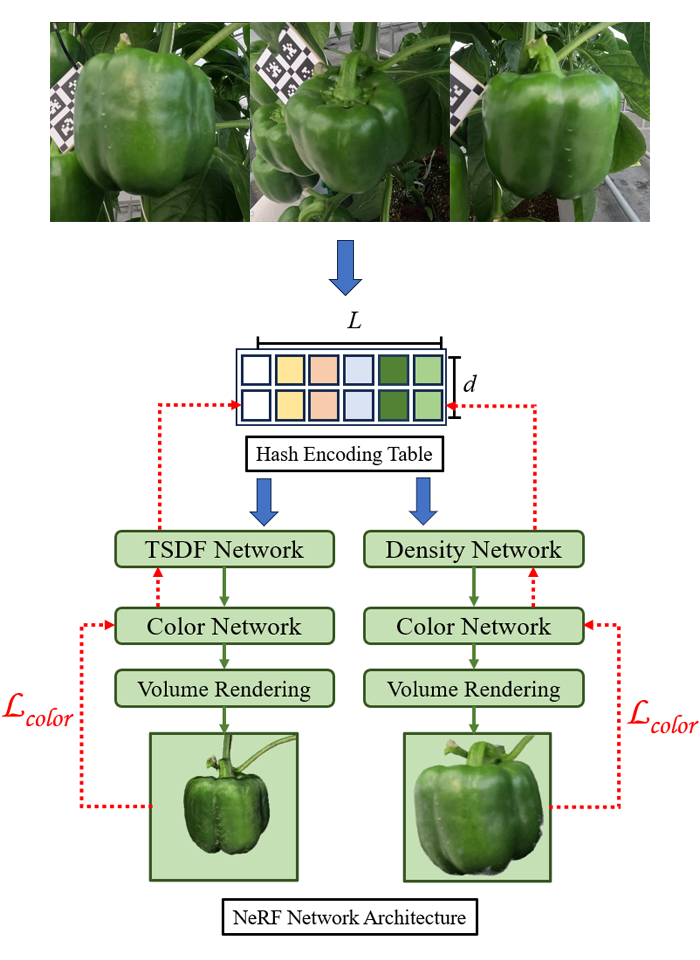}
    \caption{Neural Radiance Fields Method.}
    \label{fig:3}
\end{figure}

The Encoding is useful for extracting high-dimensional information for feature enhancement and avoiding similarity.\cite{rahaman2019spectral} Each pixel point radiates a ray and creates MLP (Multi-Layer Perception), which is trained and saves the weighting parameters. NeRF's MLP has 10 layers, with 256 neurons in each of the first 9 layers and 128 neurons in the 10th layer. Then volume rendering is used to predict the colour values of the pixel points.\cite{kajiya1984ray} The volume rendering function is:
\begin{equation}
    \label{Eq:volume rendering}
    C(\mathbf{r})=\int_{t_1}^{t_2}T(t)\cdot\sigma(\mathbf{r}(t))\cdot\mathbf{c}(\mathbf{r}(t),\mathbf{d})dt,
\end{equation}
where $T(t)$ denotes the transmittance function, which is affected by the opacity of the sampling points and the sampling distance:
\begin{equation}
    T(t)=exp(-\int_{t_n}^{t}\sigma(\mathbf{r}(s))ds).
\end{equation}
Since the sampled points are discrete, we need to take an integral of them. So we take an approximate integral to calculate the rendered color values.The approximate equation is:
\begin{equation}
    \hat{C}(\mathbf{r})=\sum^{N}_{i=1} exp(-\sum^{i-1}_{j=1}\sigma{_j}\delta{_j}) (1 - exp(-\sigma{_i}\delta{_i}))\mathbf{c}_i.
\end{equation}

During the sampling process, NeRF is divided into coarse and fine sampling to improve sampling efficiency. After equally spaced coarse sampling, the region with the greater weight is selected for fine sampling. This setting effectively reduces the extra sampling of empty points.

Finally, the loss is calculated from the rendered colour values and backpropagated:
\begin{equation}
    \mathcal{L}=\sum_{r\in R}||C(\mathbf{r})-C_{gt}(\mathbf{r})||^2_2
\end{equation}

\textbf{Neural Rendering by Instant-NGP}: Instant-NGP replaces high-dimensional location coding $\gamma(p)$ with multi-resolution hash coding\cite{muller2022instant}. The location feature values are saved in a hash table and the number of parameters can be kept in control and does not increase with the number of points.

\begin{table}[h]
    \center
    \caption{Hash Encoding Parameters}
    \renewcommand{\arraystretch}{1.2}
    \small
    \label{table:2}
    \begin{tabular}{lr}
        \toprule[2pt]
        \textbf{Parameter} & \textbf{Value} \\
        \midrule
        Number of resolutions level & 16 \\
        Hash table size & $2^{19}$\\
        Number of feature dimensions per hash table & 2 \\
        Max resolutions & 16 \\
        Min resolutions & 524288 \\
        \bottomrule[2pt]
    \end{tabular}
\end{table}
Hash Encoding in the architecture of a 3D scene, the scene is evenly divided into a voxel grid of 16 resolution levels. The coordinates of the 8 vertices of each voxel grid are fixed, and when the sampled points enter the voxel grid, they will be operated by a hash function to get the corresponding indexes and obtain the index values. The hash function is $h_i=(\tau_1x_i\oplus\tau_2y_i\oplus\tau_3z_i)\bmod T.$ where $\oplus$ denotes XOR operation, $\tau_1=1$, $\tau_2=2654435761$, and $\tau_3=805459861.$ The index values are subjected to an operation of trilinear interpolation. And accumulate the interpolated values of all resolutions, plus colour coding, as input values for the small MLP for network training.
The new encoding approach makes feature storage much more efficient, using only 5\% of NeRF's neuron count to achieve more than NeRF and increasing speed by $\times$40.

\textbf{Neural Reconstruction by Neus}: Neus is a reconstruction method that allows the surface of an object to be reconstructed more closely to the real surface. It takes the density values used by NeRF (Neural Radiance Fields) and replaces them with the point-to-surface distance (Signed Distance Field)\cite{wang2021neus}. The surface $\mathbf{S}$ of an object is the function equal to zero, the function is $\mathbf{S}=\left\{\mathrm{x}\in\mathbb{R}^3|f(\mathbf{x})=0\right\}.$
However, in 3D ray-tracing scenes, the use of the basic logistic density distribution $\phi_s(x)=se^{-sx}/(1+e^{-sx})^2.$ introduces surface biases that affect the quality of the reconstruction. Therefore, Neus introduces an unbiased and occlusion-aware weight function:
\begin{equation}
    w(t)=\frac{\phi_s(f(\mathbf{p}(t)))}{\int_0^{+\infty}\phi_s(f(\mathbf{p}(u)))\mathrm{d}u},
\end{equation}
where $\mathbf{p}(t)$ is the point on a ray from a pixel. This ensures that the final SDF value obtained is infinitely close to the surface. Neus uses essentially the same image formation model as NeRF volume rendering. 

The quality of surface reconstruction is improved by Neus, however, the training process is time-consuming and unstable.
\begin{figure*}[h]
    \centering
    \includegraphics[width=0.9\linewidth]{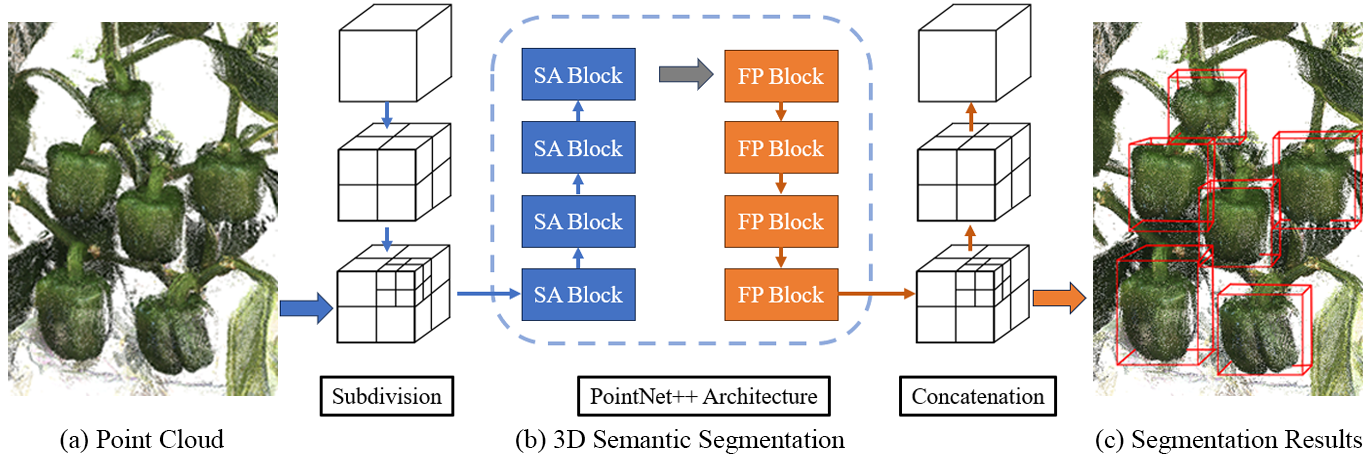}
    \caption{Illustration of our proposed 3D semantic segmentation method. (a) Input data are point cloud data extracted from 3D scanner reconstruction and NeRF reconstruction. (b) The input point cloud is segmented by the segmentation module. (c) Finally the segmentation result is obtained.}
    \label{fig:4}
\end{figure*}
Instant-NSR introduces hash coding similar to that of Instant-NGP to train at a significantly increased speed and improve the stability of network training\cite{zhao2022human}. To solve the convergence problem caused by the SDF representation in the hash coding framework, Instant-NSR introduces Truncated SDF (TSDF). The TSDF value ranges from -1 to 1. This property prevents numerical overflow of the logistic density distribution during accumulation, thereby avoiding instability during training and facilitating network convergence. The function that implements the truncated effect of TSDF is: 
\begin{equation}
    \pi(f(\mathbf{x}))=\frac{1-e^{-bf(\mathbf{x})}}{1+e^{-bf(\mathbf{x})}}.
\end{equation}
By adding constraints, it can efficiently help the model to converge and find the surface of the object.

\textbf{Our improvements on NeuS}: 
During the training phase of the Instant-NSR model, we encountered a recurring problem where voids would frequently occur during the reconstruction process. This phenomenon had a significant detrimental effect on the quality of the resulting mesh reconstruction. Various attempts were made to address this issue, including increasing the amount of image data and adjusting the camera model parameters. However, these efforts did not produce satisfactory results.
In a comparative analysis of the reconstruction results and dataset characteristics of previously studied datasets, we observed a common practice of keeping the object to be reconstructed in the centre of the scene during sampling. This practice is consistent with the ray tracing principle of NeRF. However, in the context of complex agricultural scenes, ensuring that the entire plant, along with its associated measurements, remains centred during sampling is a significant challenge.

To address this challenge, we took the initiative to reproduce the model code and subsequently modify the code responsible for calculating the scene centroid. This adaptation allowed us to set the parameters in a more appropriate range. As a result, we observed a significant improvement in the accuracy of the scene centroid calculation, leading to a significant improvement in the quality of the model reconstruction, particularly in complex agricultural scenes.

\subsection{3D modelling from Point Cloud Registration}
The Iterative Closest Point (ICP) algorithm iterates over the nearest points to obtain corresponding points. 
The point cloud data of different orientations and angles are unified under the same coordinate system through the Euler transformation relationship between corresponding points. 
This enriches the target point cloud data and makes it more complete. In 3D point cloud acquisition, the positional mapping between two sets of points is represented by displacement and rotation transformations totalling six degrees of freedom:
\begin{equation}
    \mathbf{T}^*=\underset{\mathrm{T}}{\operatorname*{\mathrm{argmax}}}\prod_ip\left(d_i^{(\mathbf{T})}\right).
\end{equation}
Suppose the optimal match is $\mathbf{T}^*$ and a pair of matching points ${a}_i$ and ${b}_i$, $\hat{b}_i=\mathbf{T}^*\hat{a}_i$. 
Define the distance variable between them as $d_i^{(\mathbf{T})}=\hat{b}_i-\mathbf{T}\hat{a}_i$.
Find the transformation matrix with the highest confidence by maximum likelihood estimation.

\subsection{Post-processing for Phenotyping Measurement}
\subsubsection{3D Pepper Detection on Point Cloud}
PointNet is a 3D segmentation method that enables efficient segmentation in natural orchards\cite{kang2023semantic}. 
The point cloud in natural orchards is inhomogeneous and disordered, lacking a structured neighbourhood region similar to an image\cite{qi2017pointnet++}. 
To address this, the PointNet network incorporates a hierarchical feature learning strategy to extract and learn features from the point cloud, which is an improvement over previous networks. 
The paper describes a hierarchical point set abstraction layer that can incrementally learn features and summarise the extracted information. 
It uses four Set Abstraction (SA) layers and Feature Propagation (FP) layers to process the point cloud at each level and performs dense prediction by propagating the point features along adjacent regions. 
The Sampling layer, Grouping layer and PointNet layer form the three-layer structure of SA. 
The sampling layer uses iterative farthest-point sampling to uniformly select a fixed number of points from the input point set. 
The grouping layer identifies neighbouring points within the local region of each centroid. 
The PointNet layer extracts features from each neighbouring region of the centroid. 
The use of continuously stacked layers in the architecture reduces the impact of undersampling on the model. 
This is achieved by allowing the model to extract and process features from multiple regions, thereby mitigating the effects of undersampling when downsampling the point cloud. 
In addition, the FP layer is used to upsample the points and propagate the processed features to each point in the ensemble.

\subsubsection{Measurement of Phenotypic data}
The 3D model reconstructed using NeRF needs to be converted to actual size for final phenotype estimation. To achieve this, scaling markers that are regular in shape and relatively complete must be found and identified. Accurate calibration plates are used as markers for the calibration reconstruction. The scale factor between the estimated and actual lengths of the calibration plate reconstructed point cloud is then used as the corrected scale of the reconstruction.
The estimation result is:
\begin{equation}
    \tau=\frac{L}{L_{\mathrm{NeRF}}},
\end{equation}
where $\tau$ is the estimated scale factor, $L$ is the actual measured length, $L_{\mathrm{NeRF}}$ is the estimated length of the scaling markers from the NeRF.

\section{Experiment and Discussion} \label{section:experiment}
\subsection{Experimental setup}
In this study, a robotic arm equipped with a high-precision 3D scanner was used to collect accurate position data for merging 3D point cloud data. Image data was simultaneously captured by a motion camera positioned at the same location. A standard agricultural dataset was used in the experiment. Initially, the robotic arm transported the sampling equipment and accurately controlled the sampling distance. The study recorded the exact position and corresponding 3D point cloud at each sampling location. The 3D point cloud data was then registered and fused, while the image data collected by the motion camera was reconstructed using the NeRF model to generate a high-fidelity reconstructed mesh. Finally, a 3D semantic segmentation network was integrated into the study to facilitate fruit phenotyping measurements.
\begin{figure}[H]
    \centering
    \includegraphics[width=1\linewidth]{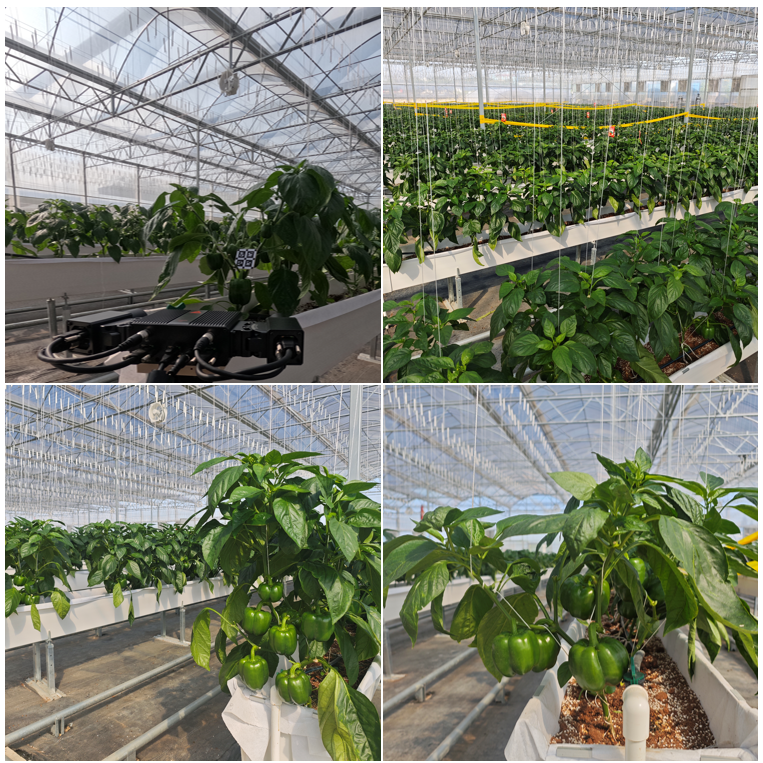}
    \caption{Illustrations of the standard scenarios for the agriculture and collection facilities.}
    \label{fig:5}
\end{figure}
\subsubsection{Dataset acquisition and processing} The study's data were obtained from the pepper planting greenhouses at Baiyun Experimental Base of Guangdong Academy of Agricultural Sciences. The acquisition methods both used a motion camera and a robotic arm-mounted 3D scanner to collect 2D image data and 3D point cloud data, respectively. Data was collected from peppers with varying growth processes and trait characteristics, including those with complex structures and occlusions, to evaluate the effectiveness of our model reconstruction.

\subsubsection{Evaluation method}
The evaluation index continues to follow the previously used PSNR(Peak Signal-to-Noise Ratio) and adds more attention to the human eye to perceive the differences in the obvious SSIM(Structure Similarity Index Measure). 
SSIM from the brightness, contrast and structure, three indicators to comprehensively assess the quality of the image, to be able to more real close to the human eye to assess the differences. 
Brightness takes the mean value, image contrast takes the standard deviation, and structural similarity takes the covariance, with the following formula:
\begin{equation}\label{Eq:ssim}
SSIM(x,y)=\frac{(2\mu_x\mu_y+C_1)(2\sigma_{xy}+C_2)}{(\mu_x^2+\mu_y^2+C_1)(\sigma_x^2+\sigma_y^2+C_2)}.
\end{equation}
\begin{equation}
    \label{Eq:PSNR}
    \mathrm{PSNR}=10\times\log_{10}\left(\frac{MAX_I^2}{\mathrm{MSE}}\right) .
\end{equation}

\subsection{Ablation Study on NeRF-based Approach}
This section presents the results of reconstructing three methods: the point cloud after alignment by a high-precision 3D scanner, Instant-NGP (the fastest NeRF model reconstruction), and Instant-NSR (the model with the best surface reconstruction quality). Additionally, it showcases the results of our optimization of the Instant-NSR model.

\begin{figure}[ht]
    \centering
    \includegraphics[width=1\linewidth]{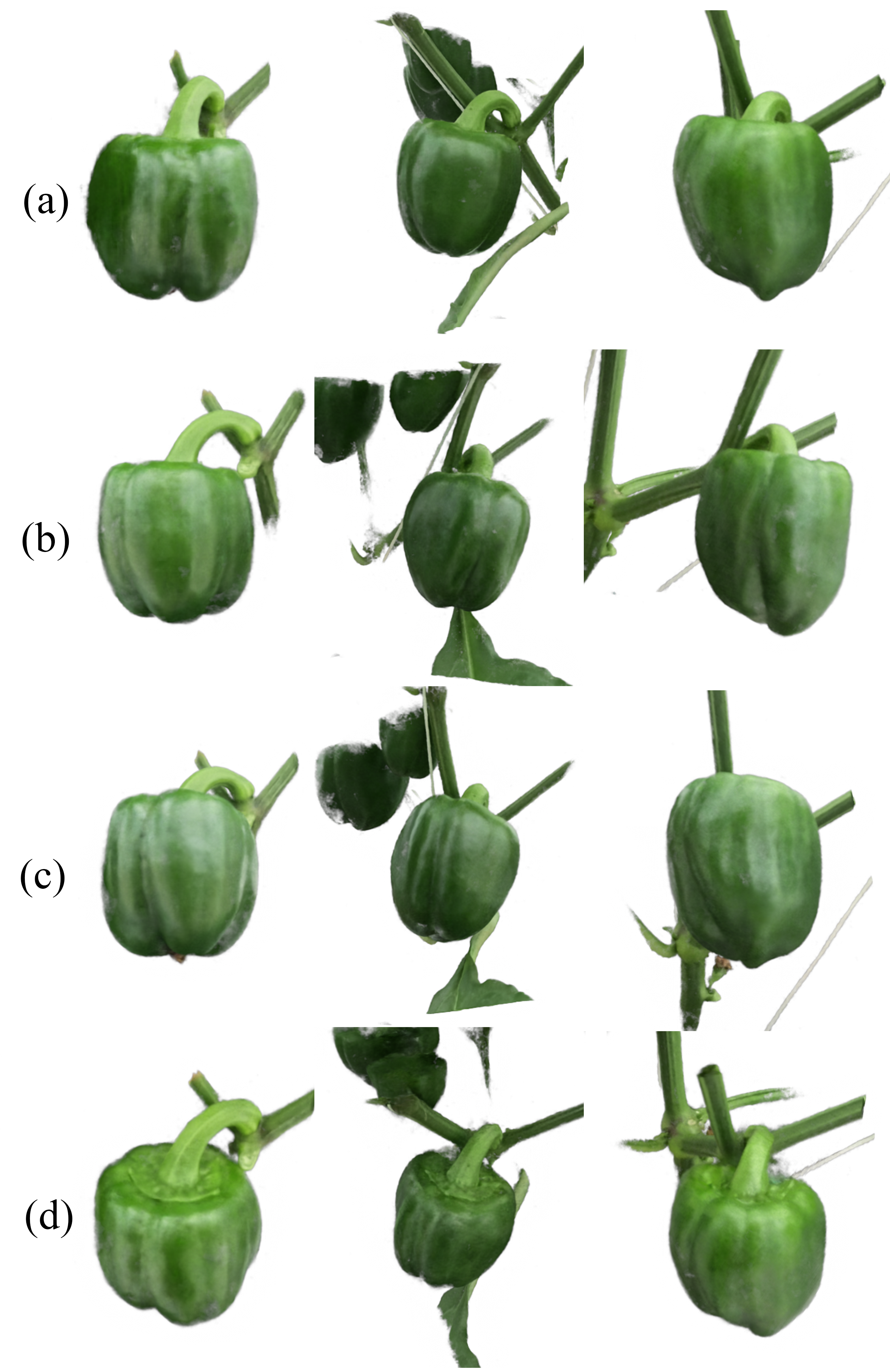}
    \caption{Multi-view renderings of the NeRF model (a) Front views (b) Side views (c) Top views (d) Elevation views.}
    \label{fig:6}
\end{figure}

Table \ref{table:3} presents a comparison of the two neural network reconstruction methods, Instant-NGP and Instant-NSR, with previous multi-view reconstruction methods. The evaluated metrics demonstrate that the Instant-NGP and Instant-NSR methods outperform the previous conventional multi-view reconstruction methods with significant time improvements. Additionally, the model's weight size is reduced, which is beneficial for model optimization and deployment. 
\begin{table}[ht]
    \center
    \caption{Average evaluation metrics of 3D reconstruction methods in the pepper dataset.}
    \renewcommand{\arraystretch}{1.4}
    \setlength{\tabcolsep}{5.4pt}
    \small
    \label{table:3}
    \begin{tabular}{lcccc}
    \toprule[2pt]
    \textbf{Methods} & \textbf{PSNR} & \textbf{SSIM} & \textbf{Train} & \textbf{Mem}  \\ \hline
    \textbf{MVS}             & 26.52 dB & 0.637 & 25min    & 245 MB  \\ 
    \textbf{Instant-NGP}     & 28.09 dB & 0.725 & 2min10s  & 34 MB   \\
    \textbf{Instant-NSR}     & 26.99 dB & 0.701 & 10min10s & 96.1 MB   \\
    \textbf{Instant-NSR(Ours)} & 28.74 dB & 0.810 & 10min36s & 95.3 MB \\ 
    \bottomrule[2pt]
    \end{tabular}
\end{table}

Instant-NGP based on density field can generate high-resolution meshes and instant renderings in a short time. However, the mesh obtained based on equal density values in a certain area lacks the accurate calculation of the surface of the reconstructed fruits due to the characteristics of the density field. As a result, modelling plants with smooth surfaces such as watermelons and bell peppers cannot achieve a smooth surface structure, which also causes the mesh colouring to deviate from the real value.
Therefore, this text highlights the advantages of using the Instant-NSR model reconstruction based on the distance sign field. Instant-NSR adds the SDF in the network, which is used to calculate the surface position. This allows the model to obtain accurate fruit surface values and export and colour the mesh. At the same time, the reconstruction of complex scenes may be slower due to the addition of new constraints. However, the model is still able to complete the reconstruction in less than 12 minutes, which is faster than previous methods and produces better results.

\begin{figure*}[]
    \centering
    \includegraphics[width=1\linewidth]{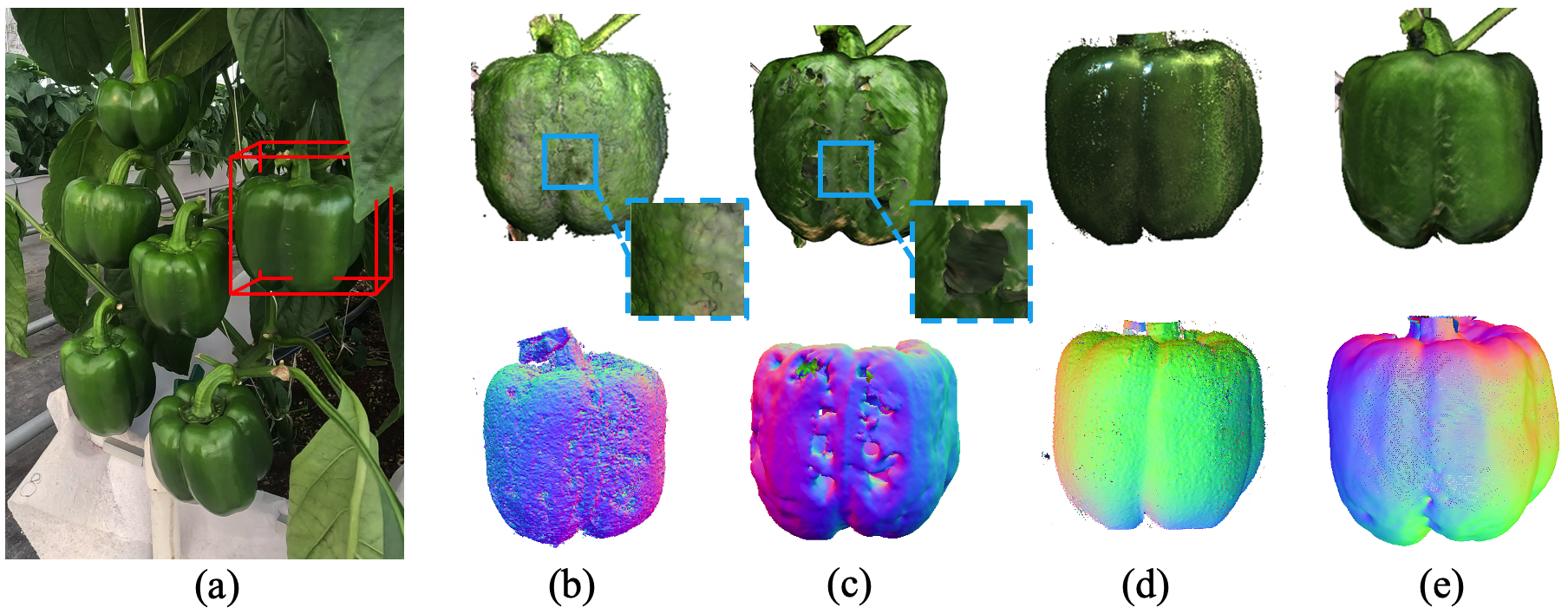}
    \caption{The figure shows a comparison between the NERF model we used and the point cloud reconstruction. The first row displays the coloured reconstruction result, while the second row shows the normals of the reconstructed model. (a) Coloured pepper environment (b) Mesh detail of Instant-NGP (c) Mesh detail of Instant-NSR before optimisation (d) Point cloud acquired by 3D scanner (e) Mesh details after Instant-NSR optimisation.}
    \label{fig:7}
\end{figure*}

An important improvement in our approach is the optimisation of the computation of the scene's centre. To demonstrate the results of this improvement, we conducted an ablation experiment.
Figures \ref{fig:7}c and e illustrate the comparison between the results obtained before and after optimising the Instant-NSR model.
The images clearly demonstrate that the optimised results are significantly improved compared to the original network reconstruction.
The coloured pepper scene reconstruction results demonstrate the effectiveness of our scene centre optimisation. 

\subsection{Quantitively Evaluation on Accuracy}
\begin{table}[H]
    \center
    \caption{Point distance compare.}
    \renewcommand{\arraystretch}{1.5}
    \setlength{\tabcolsep}{8pt}
    \small
    \label{table:4}
    \begin{tabular}{c|ccc}
        \toprule[2pt]
        \textbf{Methods} & \multicolumn{3}{c}{\textbf{Mean Distance (mm)}} \\
        \hline
        \textbf{Instant-NGP} & \textbf{1.021} & \textbf{1.131} & \textbf{0.977} \\
        \textbf{Instant-NSR} & \textbf{0.909} & \textbf{0.871} & \textbf{0.865} \\
        \bottomrule[2pt]
    \end{tabular}
\end{table}

During the reconstruction of point cloud data, it was observed that scanning objects with smooth surfaces, such as peppers, can lead to several issues with the obtained point cloud data.
(1) The object edges are often poorly reconstructed on the target edges because accurate reflections cannot be obtained from the smooth surface.
(2) Differences in colour can be observed at different angles in ambient daytime lighting, which may result in uneven colours in the reconstructed point cloud.
(3) The scanner produces many artefacts in low light.
(4) Excessive light can cause strong reflective surfaces on smooth surfaces, which can seriously affect data acquisition and create voids in the reconstruction.
All of these issues can affect the collection and measurement of plant phenotypes, resulting in inaccurate data.
Additionally, the colour of the point cloud obtained by the scanner remains fixed at the moment of acquisition. Therefore, when rendering the NeRF model, the colours are adjusted instantly according to the observation viewpoint to ensure compliance with optical rules.

\begin{figure}[ht]
    \centering
    \includegraphics[width=1\linewidth]{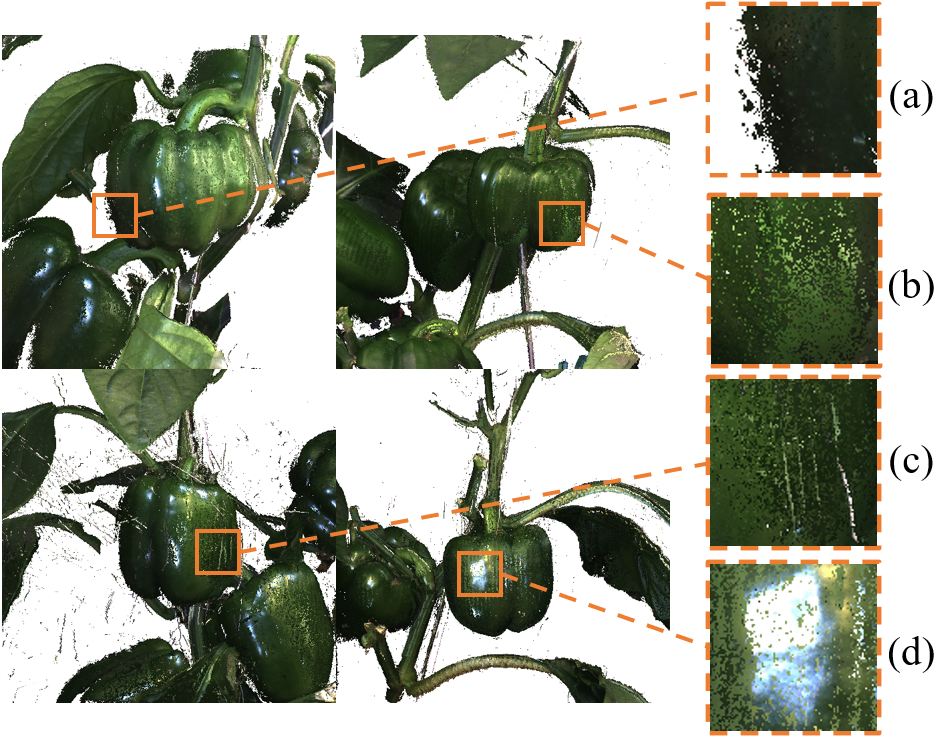}
    \caption{The figure shows the point cloud acquired by the 3D scanner and illustrates several problems with point cloud acquisition. (a) Bleeding points with cut edges on smooth surfaces. (b) Sampling chromatic aberration. (c) Insufficient light produces artefacts. (d) Reflections from highly reflective surfaces.}
    \label{fig:8}
\end{figure}

Previous studies on NeRF reconstruction often compressed the reconstructed data input with images to improve the training speed and model robustness. Although the model can maintain high resolution and accuracy after reconstruction, it loses its true scale and cannot acquire phenotypic data. To address this, we include standard calibration plates with dimensions similar to those of coloured peppers in the data acquisition process. Additionally, we added the reconstruction of the calibration plate area when reconstructing the model. Using the grid measurement tool, the reconstructed data was accurately measured. The standard length of the calibration plate reconstructed by the NeRF model was then compared with the actual value to determine the scale proportion of the model. An algorithm for scale restoration was used to restore the point cloud and mesh dimensions of the NeRF model to their true dimensions.
Finally, we compared the recovered model with the point cloud obtained from the 3D scanner and measured the average distance between the two point cloud models to assess the accuracy of the model after scale restoration. The results corresponding to the average point cloud distance for the two models used are shown in TABLE \ref{table:4}.

\begin{figure}[]
    \centering
    \includegraphics[width=0.95\linewidth]{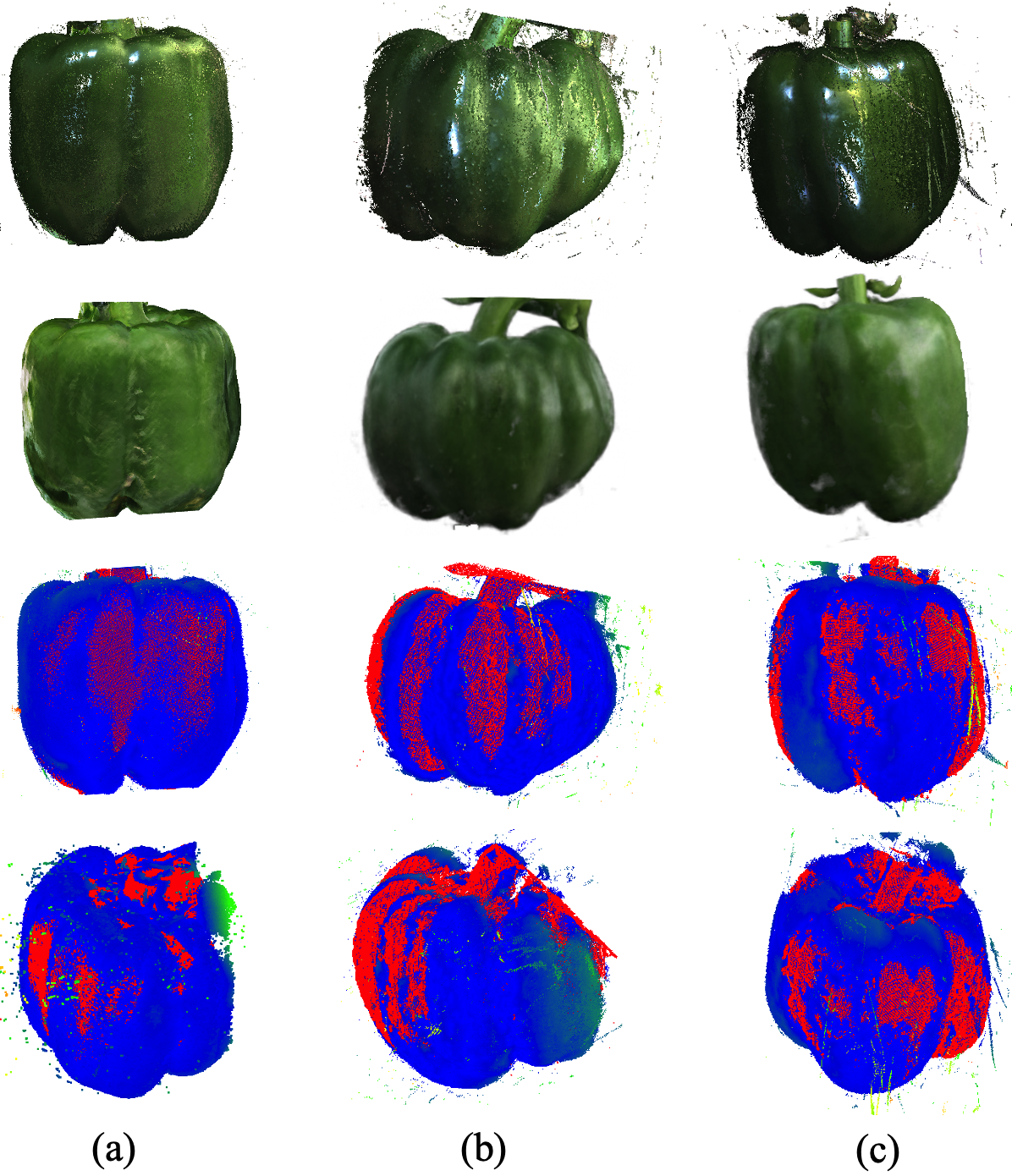}
    \caption{Illustration of matching results on three different samples by using the scanner method and instant-NSR model.}
    \label{fig:9}
\end{figure}

\subsection{Demonstration on Phenotypic Measurements}
In our previous study, we presented a deep learning-based network for 3D semantic segmentation of point clouds and post-processing. In complex agricultural scenarios, there are numerous disturbances that can affect the collection of phenotyping data, such as leaves and branches. The segmentation network is simple and efficient, enabling the model to quickly extract the fruit model for reconstruction, accurately determine the phenotyping measurement area, and perform high-precision measurements in complex scenes. 
Therefore, to ensure accurate measurements, standard calibration plates with dimensions similar to those of the coloured peppers were included in the data acquisition process. 
\begin{table}[ht]
    \center
    \caption{Comparison of phenotypic measurement data.}
   \renewcommand{\arraystretch}{1.3}
    \setlength{\tabcolsep}{5.5pt}
    \small
    \label{table:5}
    \begin{tabular}{lcccccc}
        \toprule[2pt]
        \textbf{Methods}            & \textbf{Height(cm)} & \textbf{Width(cm)} & \textbf{Difference} \\ \hline
        \textbf{Actual Measurement} & \textbf{73.0}       & \textbf{74.9}      & \textbf{}           \\
        \textbf{3D Scanner}         & \textbf{72.89}      & \textbf{75.04}     & \textbf{0.151\%}    \\
        \textbf{Instant-NGP}        & \textbf{72.78}      & \textbf{74.82}     & \textbf{0.204\%}    \\
        \textbf{Instant-NSR}        & \textbf{72.95}      & \textbf{74.89}     & \textbf{0.094\%}    \\
        \bottomrule[2pt]
    \end{tabular}
\end{table}

Table \ref{table:5} presents the results of comparing the phenotypic data obtained from the two models and 3D scanner with the true measurements. The results indicate that after the scale restoration, both Instant-NGP and Instant-NSR were able to obtain phenotypic measurements that differed from the true values by less than 1\%. Moreover, Instant-NSR achieved a higher measurement accuracy than the 3D scanner while maintaining high-quality surface results.
\begin{figure}[ht]
    \centering
    \includegraphics[width=0.95\linewidth]{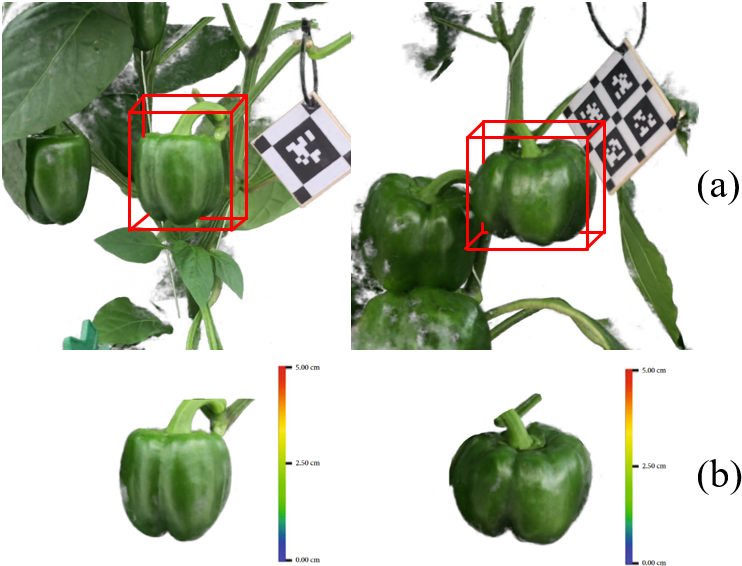}
    \caption{Illustrations of the presentation of the collected phenotyping data after 3D semantic segmentation of the point cloud. (a) 3D semantic segmentation in greenhouses pepper environment. (b) Phenotyping measurements.}
    \label{fig:10}
\end{figure}

\subsection{Discussions}
In this study, the reconstruction method for the acquisition of point clouds from 3D scanners is investigated using the Neural Radiance Fields-based 3D reconstruction method. While conventional 3D scanners are widely used for phenotypic data acquisition, the Neural Radiance Fields provide a concise and fast 3D reconstruction approach. This fast reconstruction method can overcome the difficulty of accessing heavy phenotype acquisition equipment in complex agricultural scenes. Phenotyping researchers only require lightweight camera equipment to efficiently collect phenotypes. 
\subsubsection{Model Optimisation}
Two key problems that need to be solved during the model reconstruction process are poor reconstruction quality and the existence of voids in the model body. Model parameters were optimised to ensure reconstruction quality and model training speed, resulting in better phenotypic measurement accuracy. The model's generalisation and robustness in agricultural scenarios have been improved.
\subsubsection{Reconstruction Quality Assessment}
In this experiment, we analysed the performance of two SOTA 3D reconstruction methods and a 3D scanner based on the structured light principle on the Colour Pepper dataset. We used quantitative metrics calculated from accurate point cloud distances to evaluate the performance. The results indicate that the neural network-based 3D reconstruction method can achieve similar reconstruction results to the 3D scanner in a quick and efficient manner.
\subsubsection{Scale restoration and Phenotypic measurements}
A scale restoration algorithm was incorporated into 3D reconstructed models to restore NeRF reconstructed models to their true scale. This assists in phenotypic measurements and improves the efficiency of acquiring phenotypic data from NeRF models. In addition, we have incorporated a network for 3D semantic segmentation. This allows us to extract phenotype data quickly in complex scenes and provides a path for automated detection.

\section{Conclusion and Future Work} \label{section:conclusion}
The proposed high-precision phenotype reconstruction based on the Neural Radiance Field can greatly improve the speed of phenotype acquisition. 
We have addressed the issue of the lack of true scale in the NeRF model and have simplified phenotyping measurements by comparing it with traditional acquisition methods at multiple scales. 
In the Coloured Pepper dataset, we compared the evaluation metrics of multiple methods and found that our approach yielded superior results. 
This enables future studies to obtain accurate data on plant phenotyping even after reconstruction using the NeRF model. 
Future work will focus on improving the applicability of NeRF by enabling rapid and accurate modelling in scenarios with sparse views. 
Additionally, we will explore methods to enhance background processing while preserving the reconstruction quality of the subject. 
This exploration aims to advance the application of NeRF in acquiring large-scale plant phenotyping.

\bibliographystyle{IEEEtran}
\bibliography{root}
\end{document}